\documentclass{article}
\usepackage{ijcai26}

\usepackage{times}
\usepackage{soul}
\usepackage{url}
\usepackage[hidelinks]{hyperref}
\usepackage[utf8]{inputenc}
\usepackage[small]{caption}
\usepackage{subcaption}
\usepackage{graphicx}
\usepackage{amsmath}
\usepackage{amsthm}
\usepackage{amssymb}
\usepackage{booktabs}
\usepackage{tabularx}
\usepackage{ragged2e}
\usepackage{multirow}
\usepackage[table]{xcolor}
\usepackage[ruled,linesnumbered]{algorithm2e}
\SetAlgoNlRelativeSize{0}
\SetAlCapFnt{\small}

\title{Agentic Unlearning: When LLM Agent Meets Machine Unlearning}

\author{
    Author Name
    \affiliations
    Affiliation
    \emails
    email@example.com
}
\author{
Bin Wang$^1$\and
Fan Wang$^2$\and
Pingping Wang$^1$\and
Jinyu Cong$^1$\and
Yang Yu$^4$\and
Yilong Yin$^2$\And
Zhongyi Han$^2$\thanks{Corresponding authors: hanzhongyicn@gmail.com, wbz99@sina.cn}\and
Benzheng Wei$^{1,3}$\footnotemark[1]\\
\affiliations
$^1$Center for Medical Artificial Intelligence, Shandong University of Traditional Chinese Medicine, Qingdao, China\\
$^2$School of Software, Shandong University, Jinan, China\\
$^3$School of Medical Information Engineering, Shandong University of Traditional Chinese Medicine, Jinan, China\\
$^4$Shandong Huazhi Talent Technology Co., Ltd., Jinan, China
}

\begin{document}

\maketitle

\begin{abstract}

In this paper, we introduce \textbf{agentic unlearning} which removes specified information from both model parameters and persistent memory in agents with closed-loop interaction. Existing unlearning methods target parameters alone, leaving two critical gaps: (i) parameter-memory backflow, where retrieval reactivates parametric remnants or memory artifacts reintroduce sensitive content, and (ii) the absence of a unified strategy that covers both parameter and memory pathways. 
We present Synchronized Backflow Unlearning (SBU), a framework that unlearns jointly across parameter and memory pathways. The memory pathway performs dependency closure-based unlearning that prunes isolated entities while logically invalidating shared artifacts. The parameter pathway employs stochastic reference alignment to guide model outputs toward a high-entropy prior. 
These pathways are integrated via a synchronized dual-update protocol, forming a closed-loop mechanism where memory unlearning and parametric suppression reinforce each other to prevent cross-pathway recontamination. 
Experiments on medical QA benchmarks show that SBU reduces traces of targeted private information across both pathways with limited degradation on retained data.

\end{abstract}
\section{Introduction}
\label{int}

Large Language Model (LLM) agents with persistent memory are transforming high-stakes domains such as healthcare, enabling longitudinal patient monitoring, multi-turn diagnostic reasoning, and personalized clinical support~\cite{abbasian2023conversational,shi2024ehragent,tu2025towards}. 
Their ability to write, retrieve, and update context across sessions makes them far more capable than stateless models. However, this capability introduces a critical privacy risk: sensitive information now persists in two places: model parameters and external memory stores (indices, summaries, embeddings, caches). Recent studies confirm that such dual retention leads to unintended disclosure of protected health information during interaction~\cite{carlini2021extracting,seh2020healthcare,herrera2022survey,yan2025protecting}, creating compliance challenges under HIPAA and GDPR.

\begin{figure}[t]
\centering
\includegraphics[width=0.9\columnwidth]{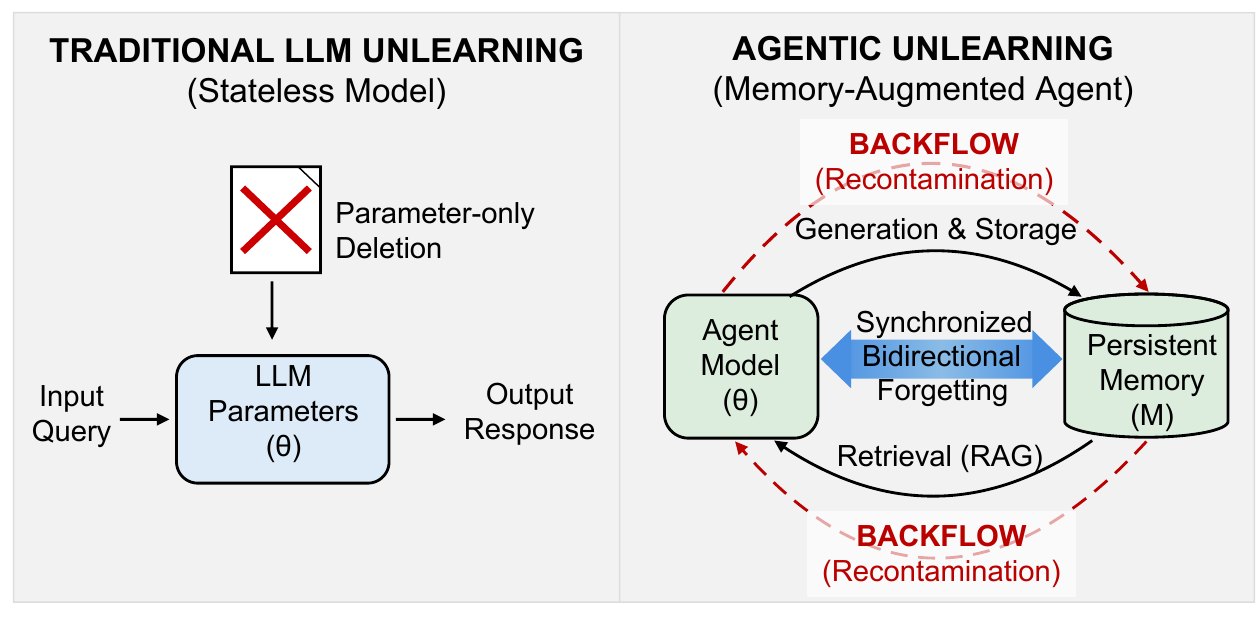}
\caption{Traditional unlearning (left) targets model parameters ($\theta$) only. Agentic unlearning (right) must address both parameters and memory to prevent backflow recontamination (red arrows). Synchronized bidirectional forgetting is required.}
\label{fig:traditional-vs-agentic}
\end{figure}

Machine unlearning provides a principled mechanism for data removal~\cite{liu2024rethinking}, yet existing LLM unlearning methods~\cite{liu2024rethinking,geng2025comprehensive,yao2024large,pawelczyk2024incontext} are designed for stateless models and focus on model-internal forgetting (either parameter updates or inference-time interventions), without addressing deletion from the persistent external memories (vector stores, summaries, interaction logs) that govern memory-augmented agents~\cite{zhong2024memorybank,packer2024memgptllmsoperatingsystems}. In memory-augmented agents, however, forgotten content persists as indices, summaries, and derived artifacts; deletion requests therefore trigger repeated recontamination through the retrieval-generation loop, a phenomenon we term \emph{backflow}. 
Even if gradient-based unlearning successfully scrubs patient data from model weights, the retrieval mechanism may still access residual traces in memory, causing the model to re-learn the forgotten information at inference time. Conversely, clearing memory alone cannot guarantee that parametric knowledge is absent, since retrieval prompts may reactivate forgotten traces encoded in the weights. This bidirectional amplification makes isolated unlearning strategies fundamentally insufficient for memory-augmented agents. This creates a backflow loop: a sensitive fact written to external memory is later retrieved into the context, influences the agent's behavior, and is then written back into new memories or re-encoded into the model during subsequent updates. Parameter unlearning alone does not break this loop; memory can reintroduce the deleted content.

Existing LLM unlearning methods, whether optimization-based~\cite{ilharco2023editing,yao2024large,jia2024soul,zhang2024negative} or prompt-based~\cite{pawelczyk2024incontext,thaker2024guardrail,liu2024large}, are insufficient for memory-augmented agents. 
Developed for stateless models, they ignore the memory hierarchy inherent to agent architectures. This hierarchy typically involves short-term memory (STM) as a transient buffer and long-term memory (LTM) for persistent storage. 
Because these methods target only parameters or ephemeral contexts, they fail to sanitize persistent memory stores and cannot prevent the backflow described above. Meanwhile, memory-oriented work focuses on retrieval augmentation rather than auditable, dependency-consistent deletion, leaving residual artifacts that enable delayed re-exposure of sensitive information. This fundamental gap motivates us to formulate \textbf{agentic unlearning}, a new paradigm that extends traditional LLM unlearning to memory-augmented agents (Figure~\ref{fig:traditional-vs-agentic}). Unlike traditional unlearning that targets only model parameters, agentic unlearning must jointly govern both parametric knowledge and persistent external memory to prevent cross-pathway recontamination. To the best of our knowledge, this work is the first to formally define and address the agentic unlearning problem for memory-augmented LLM agents.

To address these privacy challenges, we propose Synchronized Backflow Unlearning (SBU), a dual-pathway framework designed to prevent information backflow. SBU coordinates two pathways operating in tandem. The parameter pathway uses stochastic reference alignment to guide model outputs toward a high-entropy prior, suppressing implicit knowledge without catastrophic forgetting. The memory pathway performs dependency-aware deletion, using a blocklist and dependency graph to purge explicit records and their derived artifacts. These pathways are integrated via a synchronized protocol where memory unlearning is executed first. This sequence ensures the parameter update occurs on a sanitized retrieval context, preventing the model from re-encoding the information it is meant to forget. This design breaks the recontamination loop by ensuring neither the model's parameters nor its memory retains residuals capable of regenerating forgotten content. The result is a closed-loop system that provides robust and verifiable agentic unlearning. All operations are logged in a tamper-evident audit log for verifiability.

We summarize our main contributions as follows:
\begin{itemize}
    \item We are the first to define and study \textbf{agentic unlearning}, identifying its core challenge as parameter-memory backflow: a recontamination loop that renders existing, parameter-only unlearning methods ineffective.

    \item To resolve this, we propose SBU, a dual-pathway protocol that synchronizes parameter unlearning with dependency-aware memory unlearning.
    \item Experiments demonstrate that SBU prevents backflow, improving privacy by 24.8\% while maintaining $>$90\% accuracy across benchmarks.
\end{itemize}

\section{Related work}

\subsection{Machine Unlearning}
Machine unlearning has gained widespread attention~\cite{xu2023machineunlearningsurvey}, driven by emerging data privacy concerns and the pursuit of model robustness. Unlearning was first explored under partitioning data into disjoint sets to impose re-training only on the shards on which forgetting has been requested~\cite{bourtoule2021machine}. To relieve the burden of full retraining for the affected shard, a method has been proposed~\cite{neel2021descent} that achieves statistical equivalence between the post-deletion state and the state that would have existed without deletion. 
Forget-and-relearn~\cite{zhou2022fortuitous} removes undesirable features and then enforces learning good ones. 
Deviating from retraining, gradient ascent (GA) has been utilized~\cite{jang2023knowledge} instead of gradient descent to achieve targeted unlearning with only a few parameter updates. GA serves as a practical unlearning strategy in LLMs~\cite{yao2024machine}, efficiently intervening with token probabilities, making undesirable generations improbable. Incorporating well-suited loss functions and data-adaptive LoRA initializations helps resolve GA instabilities in LoRA-based unlearning~\cite{cha2025towards}.
These methods assume stateless models and target only parametric knowledge or ephemeral context, leaving the retrieval-write loop in memory-augmented agents uncontrolled. When forgotten information can be retrieved from external memory or regenerated and re-stored, parameter-only unlearning cannot prevent cross-pathway recontamination.

\subsection{Privacy Persistence in Agent Memory}
Prior agent long-term memory work optimizes retention, retrieval, and latency, while auditable forgetting remains lacking. 
MemoryBank improves persona via importance and time-weighted retention but lacks a verifiable redaction loop~\cite{zhong2024memorybank}. 
Mem0 offers a production memory layer, yet deletion and audit consistency is delegated to the application~\cite{chhikara2025mem0buildingproductionreadyai}. Virtual context and hierarchical scheduling mitigate context limits but do not govern edits or deletes of external memory~\cite{packer2024memgptllmsoperatingsystems}. Graph-augmented retrieval improves corpus-level organization, not per-user provenance and derivative-consistent deletion~\cite{edge2025localglobalgraphrag}. Multi-agent orchestration and experience replay aid collaboration and robustness but do not provide auditable forgetting~\cite{wu2024autogen,shinn2023reflexion}. Risk and memory-management studies highlight privacy and error propagation and propose utility-based add versus delete policies, but stop short of end-to-end invariants~\cite{dechant2025episodic,xiong2025memorymanagementimpactsllm}. A key survey treats forgetting as first-class and separates parameter unlearning from context deletion, but leaves integrated, traceable realizations open.

Most work pursues stronger retrieval or better organization, but lacks auditable, dependency-consistent deletion invariants. Naively deleting all derived artifacts risks destroying shared artifacts, while these methods do not coordinate with parameter-side unlearning. We propose a dual-pathway framework that enforces dependency-consistent deletion in memory while minimizing exposure in parameters.
\begin{figure*}[ht!]
\centering
\includegraphics[width=\textwidth]{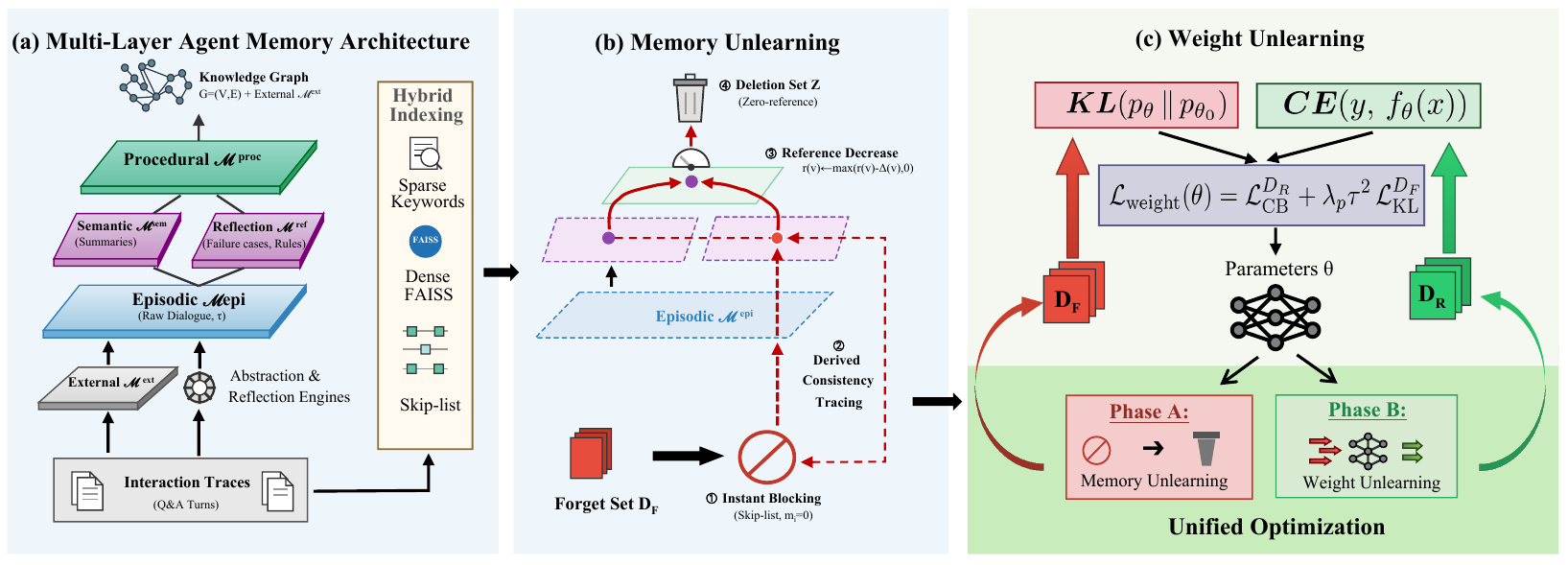}
\caption{Overview of the proposed Synchronized Backflow Unlearning (SBU) framework. The framework adopts a dual-pathway design integrating the Memory Unlearning pathway (retrieval-storage) with the Parameter Unlearning pathway (parameters).}
\label{fig:framework}
\end{figure*}
\section{Method}

We define \textbf{agentic unlearning} as the task of removing information from a memory-augmented agent. This is challenging because information is stored dually in explicit memory and implicit parameters. Deleting only explicit memory is insufficient, as the model can regenerate forgotten content from its parameters—a recontamination loop we term \emph{backflow}. We propose Synchronized Backflow Unlearning (SBU), a dual-pathway framework that prevents this by synchronizing unlearning across both representations (Figure~\ref{fig:framework}). The following subsections detail the problem formulation, memory architecture, and the two unlearning pathways.

\subsection{Agentic Unlearning}
\label{subsec:problem-formulation}
\textbf{Agentic unlearning} is the process of removing specified information from an LLM agent that uses both its internal parameters and an evolving external memory. Formally, we define an agent as $A=(\pi_{\theta}, \mathcal{M}, r, w)$, where $\pi_{\theta}$ maps interaction history to responses, $\mathcal{M}$ is the memory, and $r, w$ are functions for memory retrieval and writing. At each turn, the agent updates its memory $\mathcal{M}_{t+1}=w(\mathcal{M}_t, m_t)$, making future responses dependent on this evolving state. The objective is to transform a trained agent $A$ into an unlearned agent $A'$ that satisfies four properties: (i) it does not reveal the target information $D_{\text{tgt}}$ under adaptive interaction; (ii) its memory contains no artifacts derived from $D_{\text{tgt}}$; (iii) it is prevented from rewriting $D_{\text{tgt}}$ back into memory; while (iv) its utility on retained data is preserved.

The primary challenge that distinguishes agentic unlearning from standard LLM unlearning is a closed-loop problem we term \textbf{information backflow}. This issue arises because even after information is removed from the agent's memory ($\mathcal{M}$), residual knowledge can persist in the model's parameters ($\pi_{\theta}$). This parametric residue can then be used to regenerate the forgotten content during subsequent interactions, which is written back into memory and reverses the unlearning. Furthermore, derived artifacts in memory (e.g., summaries, knowledge graph entities) may aggregate multiple sources, requiring dependency-aware deletion to avoid destroying shared artifacts. Consequently, existing unlearning methods are insufficient as they are designed for stateless models and only target parameters, ignoring this critical memory-parameter feedback loop.

\subsubsection{Memory Architecture}
\label{subsubsec:memory-construction}

To enable propagatable and verifiable deletion, SBU models memory as a dependency graph with reference counting and blocklist enforcement. Our prototype organizes memory into multiple layers. We denote the overall memory store as
\begin{equation}
\mathcal{M}
= \mathcal{M}^{\text{epi}}
\cup \mathcal{M}^{\text{sem}}
\cup \mathcal{M}^{\text{refl}}
\cup \mathcal{M}^{\text{proc}}
\cup \mathcal{M}^{\text{ext}},
\end{equation}
where $\mathcal{M}^{\text{epi}}$ contains episodic dialogue traces (formalized as $M$), $\mathcal{M}^{\text{sem}}$ stores semantic summaries ($S$), and $\mathcal{M}^{\text{refl}}$ stores reflections ($R$). Our unlearning guarantees depend on the graph structure over $\{M, S, R, K\}$, not on the specific content of any memory layer. Full implementation details are provided in the supplementary material.

As illustrated in Figure~\ref{fig:framework}a, each memory is represented as a node in a dependency graph $G=(V,E)$, where nodes $v\in V$ include raw memories, derived summaries, reflections, and knowledge graph entities; edges $E$ encode derivation relationships.
Each node maintains a reference counter $r(v)$ tracking how many nodes depend on it. A persistent blocklist $B$ stores identifiers of deleted memories, enabling $O(1)$ membership checks to prevent re-exposure. Memory contents are indexed via hybrid search combining symbolic keyword matching and dense vector retrieval, with the blocklist enforced at retrieval boundaries. This provenance-aware representation enables the memory unlearning pathway to propagate deletions through dependency chains while preserving shared artifacts.

\subsection{Synchronized Backflow Unlearning (SBU)}
\subsubsection{Memory Unlearning}

To address deleting memories without destroying shared artifacts, we introduce a dependency-aware unlearning pathway. Derived artifacts aggregate multiple sources; naively deleting descendants would break shared artifacts. Our approach prunes artifacts supported exclusively by forgotten data while preserving those with remaining valid sources. Implementation details on data structures and cryptographic verification are provided in the supplementary material.

\indent\textit{Formalization.} Let $M$ denote the set of raw episodic memories, $S$ the set of semantic summaries, $R$ the set of reflections, and $K$ the set of knowledge graph nodes. We define a dependency graph $G=(V,E)$ over the vertex set $V = M \cup S \cup R \cup K$. For a deletion request $D_F \subseteq M$, we define the dependency closure as
\begin{equation}
\small
\mathrm{Dep}(D_F) = \{ v \in (S \cup R \cup K) \mid \exists m \in D_F \text{ such that } m \leadsto v \text{ in } G \},
\end{equation}
where $m \leadsto v$ denotes reachability in the dependency graph. The unlearning operation updates the blocked set $B' = B \cup D_F$ and excludes $B'$ from retrieval, then removes $M' = M \setminus D_F$, $S' = S \setminus \mathrm{Dep}(D_F)$, $R' = R \setminus \mathrm{Dep}(D_F)$, $K' = K \setminus \mathrm{Dep}(D_F)$, with postcondition $D_F \cap M' = \emptyset$ and $\mathrm{Dep}(D_F) \cap (S' \cup R' \cup K') = \emptyset$.

To realize the dependency-aware deletion described above, Figure~\ref{fig:framework}b illustrates the unified memory unlearning pipeline executed on each forget request for target memories $D_F \subseteq M$. First, target memory IDs are immediately added to a persistent blocked set $B \leftarrow B \cup D_F$, enabling $O(1)$ hash-set membership checks per candidate during retrieval. Then, the system traverses the dependency graph from target memories to derived artifacts (reflections, summaries, KG nodes), using reference counting to distinguish exclusively-dependent artifacts from shared ones. Reflections are marked as outdated, reference counts are decremented for shared entities, and zero-reference nodes are batch-removed, ensuring that shared artifacts depending on retained memories are preserved. Finally, the target memories are deleted from storage along with their vector representations. To control vector-index staleness, the system periodically rebuilds the vector index when $|B|$ exceeds threshold $\tau=100$.

\indent\textit{Complexity.} Retrieval incurs $O(k \cdot r)$ filtering overhead beyond base ANN search, where $k$ is the top-$k$ parameter and $r \approx 2\text{--}3$ is the oversampling factor to maintain result quality. Offline vector-index reconstruction has $O(N \cdot d)$ complexity ($N$ active memories, $d{=}1536$ embedding dimension), amortized to $O(N \cdot d / \tau)$ per deletion. Cleanup adds $O(|V_{\text{vis}}| + |E_{\text{vis}}|)$ graph traversal cost for visited nodes and edges in the dependency subgraph.

\indent\textit{Consistency.} The pathway maintains two invariants. \textit{Invariant 1 (Blocking completeness):} For retrieval paths consulting the blocked set, no memory $m \in B$ appears in results. \textit{Invariant 2 (Dependency consistency):} Derived artifacts depending on deleted memories are marked as outdated or have their reference counts decremented accordingly. All deletion operations are logged in a tamper-evident write-ahead log with hash-chain verification; see the supplementary material for details.

\subsubsection{Parameter Unlearning}

To prevent parametric recontamination where residual weights regenerate forgotten content, we introduce an Entropy-Regularized Parameter Unlearning pathway. Memory deletion alone cannot prevent backflow because the base model can regenerate forgotten content and the agent can re-store it as new memory; parameter unlearning closes this loop by making the model's distribution on forget queries intentionally non-informative. GA maximizes loss on forget data, but tends to produce incorrect predictions and causes large parameter drift that damages retain performance. NPO adjusts relative preferences between outputs, but is designed for preference tuning rather than complete knowledge removal. Our approach instead aligns the model's output distribution on forget queries to a high-entropy prior, making the model maximally uncertain rather than confidently wrong, which better preserves critical medical knowledge on the retain set.

To realize this parameter-level deletion, Figure~\ref{fig:framework}c illustrates the parameter unlearning pathway, which operates as a KL-to-random scheme.
Instead of performing gradient ascent on the forget set, we introduce a frozen reference model $f_{\theta_0}$ that is randomly initialized and encourage the current model $f_\theta$ to match this random-like distribution on the forget set, while preserving performance on the retain set.
Let $D_F$ and $D_R$ denote the forget and retain sets, and let $p_\theta(\cdot\mid x)$ and $p_{\theta_0}(\cdot\mid x)$ be the output token distributions of the student and reference models with an optional temperature $T$. The parameter-level objective is:
\begin{equation}
L_{\text{weight}}(\theta) = L_{\text{CE}}^{D_R} + \lambda_F T^2 L_{\text{KL}}^{D_F},
\end{equation}
where $L_{\text{CE}}^{D_R} = \mathbb{E}_{(x,y)\in D_R}[\mathrm{CE}(y,f_\theta(x))]$ is the cross-entropy loss on the retain set, and $L_{\text{KL}}^{D_F} = \mathbb{E}_{x\in D_F}[\mathrm{KL}(p_\theta \,\Vert\, p_{\theta_0})]$ is the KL divergence on the forget set; here $\lambda_F>0$ balances retention and forgetting, and
\begin{equation}
p_\theta=\mathrm{softmax}(z_\theta/T),\quad
p_{\theta_0}=\mathrm{softmax}(z_{\theta_0}/T),
\end{equation}
with $z_\theta$ and $z_{\theta_0}$ denoting the pre-softmax logits.
On retain samples, the model is trained with standard cross-entropy to maintain utility, whereas on forget samples, the KL term drives the output distribution towards that of a randomly initialized model, effectively erasing fine-grained information while keeping the logits in a high-entropy regime.
In practice, we implement this objective in a mixed-batch trainer: each mini-batch contains both retain and forget samples tagged with a factor flag, and the total loss is computed as a weighted sum of cross-entropy (for retain samples) and KL divergence (for forget samples).
An alternative approach would be to directly maximize the entropy of the output distribution on the forget set. However, our KL-to-random scheme provides a more stable learning target by aligning to a structured high-entropy prior from a reference model, rather than forcing the output towards a perfectly uniform distribution, which can risk over-unlearning and damaging model capabilities.

\subsubsection{Unified Optimization}

SBU coordinates memory and parameter updates to prevent recontamination. Given a deletion request $D_F$, we execute both pathways sequentially: (1) block and remove target data from retrieval, then (2) update model parameters to suppress the deleted content.
The memory pathway first adds the request to the blocklist, $B \leftarrow B \cup D_F$, to prevent further retrieval exposure, then removes items in the dependency closure $\mathrm{Dep}(D_F)$ to eliminate derived artifacts. The parameter pathway then minimizes $L_{\text{weight}}$ to suppress parametric dependence on the deleted content. This order ensures that parameter optimization operates on a clean retrieval context, preventing gradients from re-encoding the target. The process repeats for incremental requests, with periodic index compaction when $|B| > \tau$.

\begin{algorithm}[t]
\small
\caption{SBU}
\label{alg:SBU}
\KwIn{Forget set $D_F$, retain set $D_R$, model $f_\theta$, reference $f_{\theta_0}$, temperature $T$, coefficient $\lambda_F$, iterations $T_{\text{max}}$.}
\KwOut{Updated model $\theta^*$ and memory system.}
\vspace{1mm}
\textbf{for} $t = 1$ \textbf{to} $T_{\text{max}}$ \textbf{do}\\[2pt]
\quad\textbf{A. Memory Unlearning:}\\
\quad Block targets: $B \leftarrow B \cup D_F$.\\
\quad Prune dependency closure $C \leftarrow \mathrm{Dep}(D_F)$.\\
\quad Delete $D_F$ and vectors.\\
\quad Rebuild index if $|B| > \tau$.\\
\quad Archive logs.\\[2pt]
\quad\textbf{B. Parameter Unlearning:}\\
\quad \textbf{for} each mini-batch $(B_R, B_F)$ from $(D_R, D_F)$ \textbf{do}\\
\quad \quad Compute $L_{\text{CE}} = \frac{1}{|B_R|}\sum_{(x,y)\in B_R}\mathrm{CE}(y, f_\theta(x))$.\\
\quad \quad Compute $L_{\text{KL}} = \frac{1}{|B_F|}\sum_{x\in B_F}\mathrm{KL}(p_\theta \,\Vert\, p_{\theta_0})$.\\
\quad \quad Update $\theta \leftarrow \theta - \eta \nabla_\theta (L_{\text{CE}} + \lambda_F T^2 L_{\text{KL}})$.\\
\quad \textbf{end for}\\
\textbf{end for}\\[2pt]
\textbf{C. Output:}\\
Output updated $\theta^*$, cleaned memory, and audit records.
\end{algorithm}

\section{Experiment}
\subsection{Setup}
\indent\textit{\textbf{Dataset Summary.}}
We evaluate on three medical QA benchmarks: (1) MedMCQA~\cite{pmlr-v174-pal22a} contains ${\sim}$183k multiple-choice questions from AIIMS/NEET-PG entrance exams, spanning 2,400 topics across 21 medical subjects; (2) MedQA~\cite{jin2021disease} contains ${\sim}$10k multiple-choice questions from professional board exams; (3) MedReason~\cite{wu2025medreasonelicitingfactualmedical} contains ${\sim}$33k question-answer pairs from seven datasets targeting open-ended generation; we use only the pairs and exclude reasoning annotations. We conduct experiments with two forget set sizes: QF=100 and QF=1000.

\indent\textit{\textbf{Evaluation Metrics.}}
We evaluate unlearning with four metrics. (1) Accuracy on the Forget Set measures removal of targeted knowledge; lower indicates successful forgetting. (2) Accuracy on the Test Set measures generalization to unseen samples from the same distribution. 
(3) Generalization (Gen.) measures retained capability on held-out QA benchmarks; higher accuracy indicates better preservation of general medical knowledge. 
(4) Membership Inference Attack (MIA) score assesses unlearning from a privacy perspective. We compute the area under the ROC curve \( \mathcal{A} \) from loss distributions of member vs. non-member data: \( \mathcal{A}\approx0.5 \) is ideal, values near 1 indicate under-unlearning, near 0 indicate over-unlearning. We normalize \( \mathrm{MIA}=1-2|\mathcal{A}-0.5| \in [0,1] \).

\indent\textit{\textbf{Baselines.}} 
We compare two categories of baselines. Parameter-level baselines include: (1) Gradient Ascent (GA), which optimizes $-\mathcal{L}_{CE}$ on the forget set; (2) NPO~\cite{zhang2024negative}, which uses negative preference optimization; (3) Sequential LoRA~\cite{premptis2025ails}/Retrain, which fine-tunes with data chunking; (4) Adapter Merging~\cite{xu2025zjuklab}, which merges adapters via TIES; (5) Original Model as reference. Memory-side baselines fix LLM parameters and only modify memory: (1) Naive Deletion removes target entries; (2) Re-indexing rebuilds vector indices; (3) Retraining Oracle reconstructs memory from the retain set.

\indent\textit{\textbf{Unlearning model.}} 
We use II-Medical-8B~\cite{2025II-Medical-8B}, a medical LLM built on Qwen3-8B and fine-tuned on medical QA datasets to enhance domain-specific reasoning. The integration of external memory improves performance across all benchmarks, as shown in Table~\ref{tab:memory-mechanism}.
\begin{table}[h!]
\centering
\footnotesize
\caption{Accuracy (\%) comparison of II-Medical-8B with and without memory mechanism.}
\label{tab:memory-mechanism}
\setlength{\tabcolsep}{3pt}
\begin{tabular}{@{}lcccc@{}}
\toprule
Model & MedQA $\uparrow$ & MedMCQA $\uparrow$ & MedReason $\uparrow$ & Avg. $\uparrow$ \\
\midrule
II-Medical-8B & 78.39 & 84.71 & 70.00 & 77.70 \\
\rowcolor{gray!10}
\textbf{+ Memory} & \textbf{88.00} & \textbf{87.67} & \textbf{87.67} & \textbf{87.78} \\
\bottomrule
\end{tabular}
\end{table}

\indent\textit{\textbf{Implementation Details.}} 
We use OpenAI's text-embedding-ada-002 to encode memories into 1536-dimensional vectors. The memory system stores 2000 entries and retrieves top-5 via hybrid search, combining semantic similarity (weight 0.7) and keyword matching (weight 0.3). We report mean and standard deviation over 3 runs, with best in \textbf{bold} and second-best \underline{underlined}.

\subsection{Main Results}
\label{subsec:main-results}

\begin{table*}[ht!]
\centering
\small
\caption{Performance comparison of different unlearning methods on MedQA (QF=100)}
\label{tab:medqa-qf100}
\resizebox{0.95\textwidth}{!}{%
\begin{tabular}{@{}l|cc|c|cc@{}}
\toprule
\multirow{2}{*}{Method} &
\multicolumn{2}{c|}{Accuracy (\%)}&
Domain (\%) &
\multicolumn{2}{c}{Privacy Metrics} \\
\cmidrule(lr){2-3} \cmidrule(lr){4-4} \cmidrule(lr){5-6}
& Forget $\downarrow$ & Test $\uparrow$ & Gen. $\uparrow$ & MIA AUC $\rightarrow$0.5 & MIA Score $\uparrow$ \\
\midrule
Sequential LoRA~\cite{premptis2025ails} &
$89.33_{\pm 2.31}$ &
$88.67_{\pm 1.15}$ &
$87.00_{\pm 2.65}$ &
$0.642_{\pm 0.0004}$ &
$0.7167_{\pm 0.0081}$ \\
Sequential Retrain~\cite{premptis2025ails} &
$91.33_{\pm 3.06}$ &
$\underline{89.67_{\pm 4.18}}$ &
$88.00_{\pm 3.47}$ &
$0.636_{\pm 0.0006}$ &
$0.7287_{\pm 0.0110}$ \\
GA &
$\underline{89.00_{\pm 4.24}}$ &
$87.00_{\pm 4.24}$ &
$\underline{89.50_{\pm 2.12}}$ &
$\underline{0.618_{\pm 0.038}}$ &
$\underline{0.7640_{\pm 0.0750}}$ \\
NPO~\cite{zhang2024negative} &
$90.00_{\pm 5.29}$ &
$87.67_{\pm 6.05}$ &
$88.33_{\pm 2.31}$ &
$0.637_{\pm 0.0006}$ &
$0.7267_{\pm 0.0110}$ \\
Adapter Merging~\cite{xu2025zjuklab} &
$91.33_{\pm 1.15}$ &
$88.00_{\pm 3.65}$ &
$89.34_{\pm 0.58}$ &
$0.636_{\pm 0.0006}$ &
$0.7287_{\pm 0.0110}$ \\
Origin &
$89.33_{\pm 2.04}$ &
$91.00_{\pm 3.54}$ &
$87.33_{\pm 3.09}$ &
$0.637_{\pm 0.000}$ &
$0.7267_{\pm 0.0110}$ \\
\rowcolor{gray!10}
\textbf{Ours} &
$\mathbf{73.00_{\pm 1.41}}$ &
$\mathbf{92.50_{\pm 2.12}}$ &
$\mathbf{90.50_{\pm 0.71}}$ &
$\mathbf{0.552_{\pm 0.0009}}$ &
$\mathbf{0.8953_{\pm 0.0181}}$ \\
\bottomrule
\end{tabular}%
}
\end{table*}
\begin{table*}[ht!]
\centering
\small
\caption{Performance comparison of different unlearning methods on MedMCQA (QF=1000)}
\label{tab:medmcqa-qf1000}
\resizebox{0.95\textwidth}{!}{%
\begin{tabular}{@{}l|cc|c|cc@{}}
\toprule
\multirow{2}{*}{Method} &
\multicolumn{2}{c|}{Accuracy (\%)} &
Domain (\%) &
\multicolumn{2}{c}{Privacy Metrics} \\
\cmidrule(lr){2-3} \cmidrule(lr){4-4} \cmidrule(lr){5-6}
& Forget $\downarrow$ & Test $\uparrow$ & Gen. $\uparrow$ & MIA AUC $\rightarrow$0.5 & MIA Score $\uparrow$ \\
\midrule
Sequential LoRA~\cite{premptis2025ails} &
$\underline{76.67_{\pm 7.93}}$ &
$87.67_{\pm 4.88}$ &
$\underline{89.00_{\pm 2.83}}$ &
$\underline{0.5080_{\pm 0.0071}}$ &
$\underline{0.9840_{\pm 0.0141}}$ \\
Sequential Retrain~\cite{premptis2025ails} &
$90.00_{\pm 2.00}$ &
$88.00_{\pm 2.55}$ &
$88.67_{\pm 1.85}$ &
$0.5090_{\pm 0.0057}$ &
$0.9820_{\pm 0.0111}$ \\
GA &
$87.33_{\pm 3.06}$ &
$\underline{88.67_{\pm 2.53}}$ &
$87.67_{\pm 1.77}$ &
$\underline{0.5080_{\pm 0.0200}}$ &
$\underline{0.9840_{\pm 0.0040}}$ \\
NPO~\cite{zhang2024negative} &
$86.00_{\pm 8.72}$ &
$87.00_{\pm 4.40}$ &
$83.00_{\pm 5.50}$ &
$0.5090_{\pm 0.0458}$ &
$0.9820_{\pm 0.0092}$ \\
Adapter Merging~\cite{xu2025zjuklab} &
$88.67_{\pm 3.06}$ &
$\underline{88.67_{\pm 1.72}}$ &
$85.33_{\pm 3.88}$ &
$0.5087_{\pm 0.0513}$ &
$0.9827_{\pm 0.0103}$ \\
Origin &
$89.00_{\pm 1.41}$ &
$88.00_{\pm 0.71}$ &
$\mathbf{90.00_{\pm 0.71}}$ &
$0.5490_{\pm 0.0180}$ &
$0.9020_{\pm 0.0361}$ \\
\rowcolor{gray!10}
\textbf{Ours} &
$\mathbf{71.33_{\pm 7.02}}$ &
$\mathbf{89.33_{\pm 1.53}}$ &
$\underline{89.00_{\pm 1.00}}$ &
$\mathbf{0.5000_{\pm 0.0265}}$ &
$\mathbf{0.9960_{\pm 0.0020}}$ \\
\bottomrule
\end{tabular}%
}
\end{table*}

\indent\textit{\textbf{Results on Medical QA Benchmarks.}}
Table~\ref{tab:medqa-qf100} presents the primary results on MedQA (QF=100). Conventional parameter-efficient baselines exhibit a critical vulnerability: while Sequential LoRA preserves utility (test: 88.67\%), it fails to mask data membership, yielding a MIA Score of 0.717, scarcely better than the Original model (0.727). In contrast, SBU achieves a MIA Score of 0.895, representing a 24.8\% improvement in privacy protection, while maintaining test accuracy (92.50\%) and generalization (90.50\%). On MedMCQA (QF=100, Supplementary Table 2), SBU achieves test/gen of 92.33\%/92.00\% with a MIA Score of 0.973. On MedReason (QF=100, Supplementary Table 3), SBU reaches test/gen of 87.00\%/89.00\% with MIA Score of 0.891. In contrast, methods that optimize forget aggressively exhibit catastrophic over-unlearning. For example, NPO attains the lowest forget (74.00\%) but suffers severe generalization collapse (gen: 41.67\%), highlighting that surgical unlearning requires preserving capability while removing membership signals.
This is because baselines only modify LLM parameters while leaving the memory bank unchanged, allowing privacy leakage to persist through retrieval.

\indent\textit{\textbf{Scalability and Resilience.}}
When the forget set increases to 1000 (Supplementary Table 1), baseline privacy metrics stagnate (MIA Score $\approx 0.62$), whereas SBU achieves 0.802 while maintaining test/gen at 90.83\%/89.67\%. On MedMCQA (QF=1000, Table~\ref{tab:medmcqa-qf1000}), SBU reaches a MIA Score of 0.996. On MedReason (QF=1000, Supplementary Table 4), SBU achieves a MIA Score of 0.990 while preserving gen at 89.80\%, whereas NPO collapses to gen 62.33\%.

\begin{figure}[ht!]
\centering
\includegraphics[width=\columnwidth]{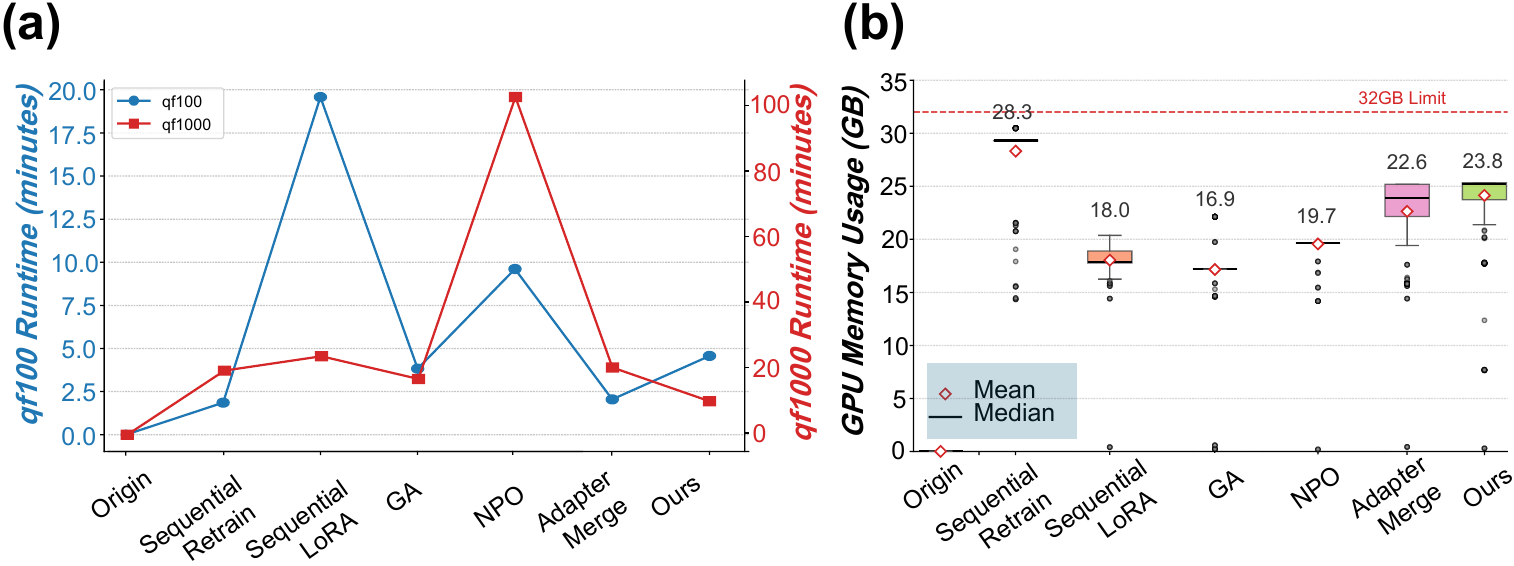}
\caption{Computational efficiency. (a) Runtime vs. forget set size for QF100 and QF1000. (b) GPU memory usage during training. Red diamonds indicate the mean; dashed line marks device capacity.}
\label{fig:efficiency}
\end{figure}

\begin{figure}[ht!]
\centering
\includegraphics[width=\columnwidth]{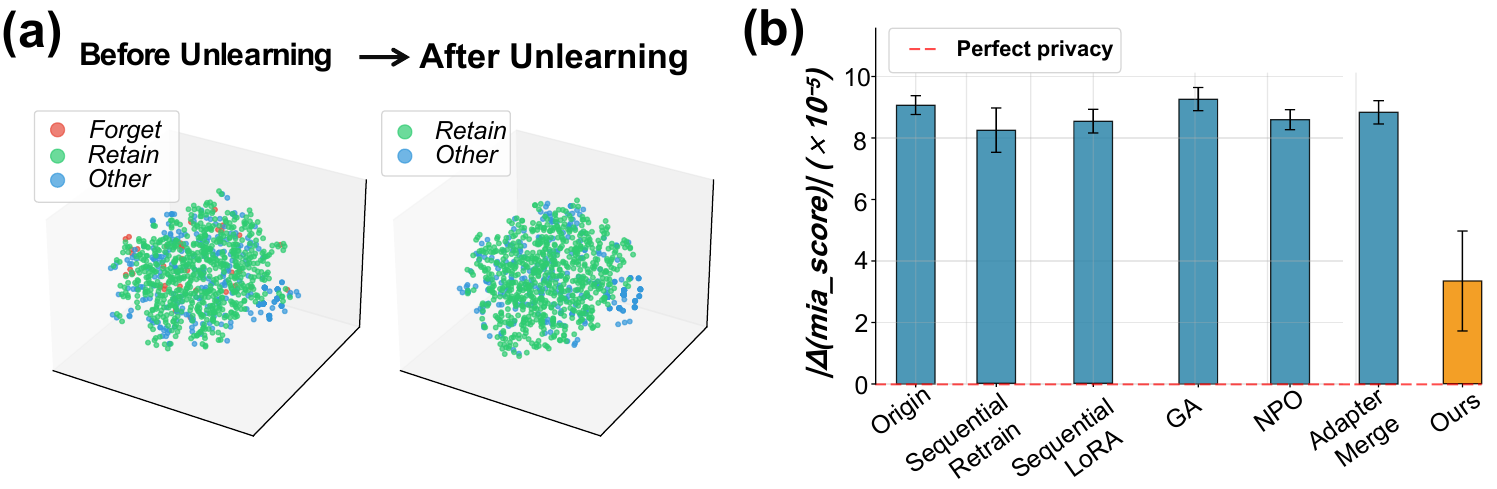}
\caption{Privacy and memory analysis. (a) Memory embeddings before and after unlearning. (b) Privacy metric $|\Delta(\text{MIA\_score})|$ ($\times 10^{-5}$); lower is better.}
\label{fig:mia-memory}
\end{figure}

\indent\textit{\textbf{Efficiency and Privacy Analysis.}}
We evaluate the computational overhead of our method in Figure~\ref{fig:efficiency}. SBU maintains lower GPU memory usage compared to baselines (Figure~\ref{fig:efficiency}b) and scales well as the forget set size increases (Figure~\ref{fig:efficiency}a). We further validate the effectiveness of privacy erasure in Figure~\ref{fig:mia-memory}. The MIA scores show that SBU achieves minimal divergence between member and non-member distributions, confirming that the unlearning process effectively eliminates distinguishable membership traces from the model's output across both pathways.

\indent\textit{\textbf{Memory-side Forgetting Analysis.}}
We examine the effect of our memory unlearning pathway on the external memory system. As shown in Table~\ref{tab:memory-before-after}, for MedQA with QF=100, the memory accuracy on the forget set drops from 78\% to 14\% after unlearning, while the memory accuracy on the retain set slightly increases from 54\% to 56\%. This indicates that the memory pathway effectively suppresses exposure of forgotten samples without harming retrieval quality on retained knowledge. Figure~\ref{fig:mia-memory}b visualizes this effect, showing that forget-related memories are removed while retained memory geometry remains largely unchanged.

\begin{table}[ht!]
\centering
\footnotesize
\caption{Memory accuracy (\%) before and after memory unlearning.}
\label{tab:memory-before-after}
\setlength{\tabcolsep}{3pt}
\begin{tabular}{@{}lcccccc@{}}
\toprule
& \multicolumn{2}{c}{MedQA} & \multicolumn{2}{c}{MedMCQA} & \multicolumn{2}{c}{MedReason} \\
\cmidrule(lr){2-3} \cmidrule(lr){4-5} \cmidrule(lr){6-7}
& Forget $\downarrow$ & Retain $\uparrow$ & Forget $\downarrow$ & Retain $\uparrow$ & Forget $\downarrow$ & Retain $\uparrow$ \\
\midrule
Before & 78.0 & 54.0 & 84.0 & \textbf{92.0} & 68.0 & \textbf{64.0} \\
\rowcolor{gray!10}
\textbf{After} & \textbf{14.0} & \textbf{56.0} & \textbf{30.0} & 90.0 & \textbf{26.0} & \textbf{64.0} \\
\bottomrule
\end{tabular}
\end{table}

\begin{figure}[ht!]
\centering
\includegraphics[width=\columnwidth]{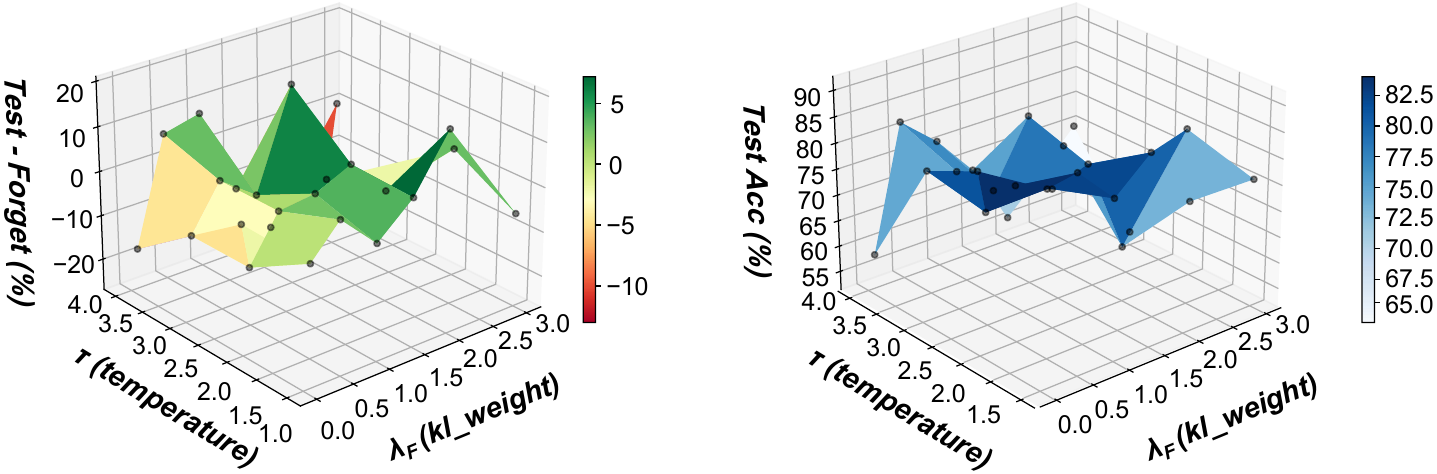}
\caption{Visualization of hyperparameter sensitivity, showing the interaction between $\lambda_F$ and $T$.}
\label{fig:3d-surface}
\end{figure}

\begin{figure}[ht!]
\centering
\includegraphics[width=\columnwidth]{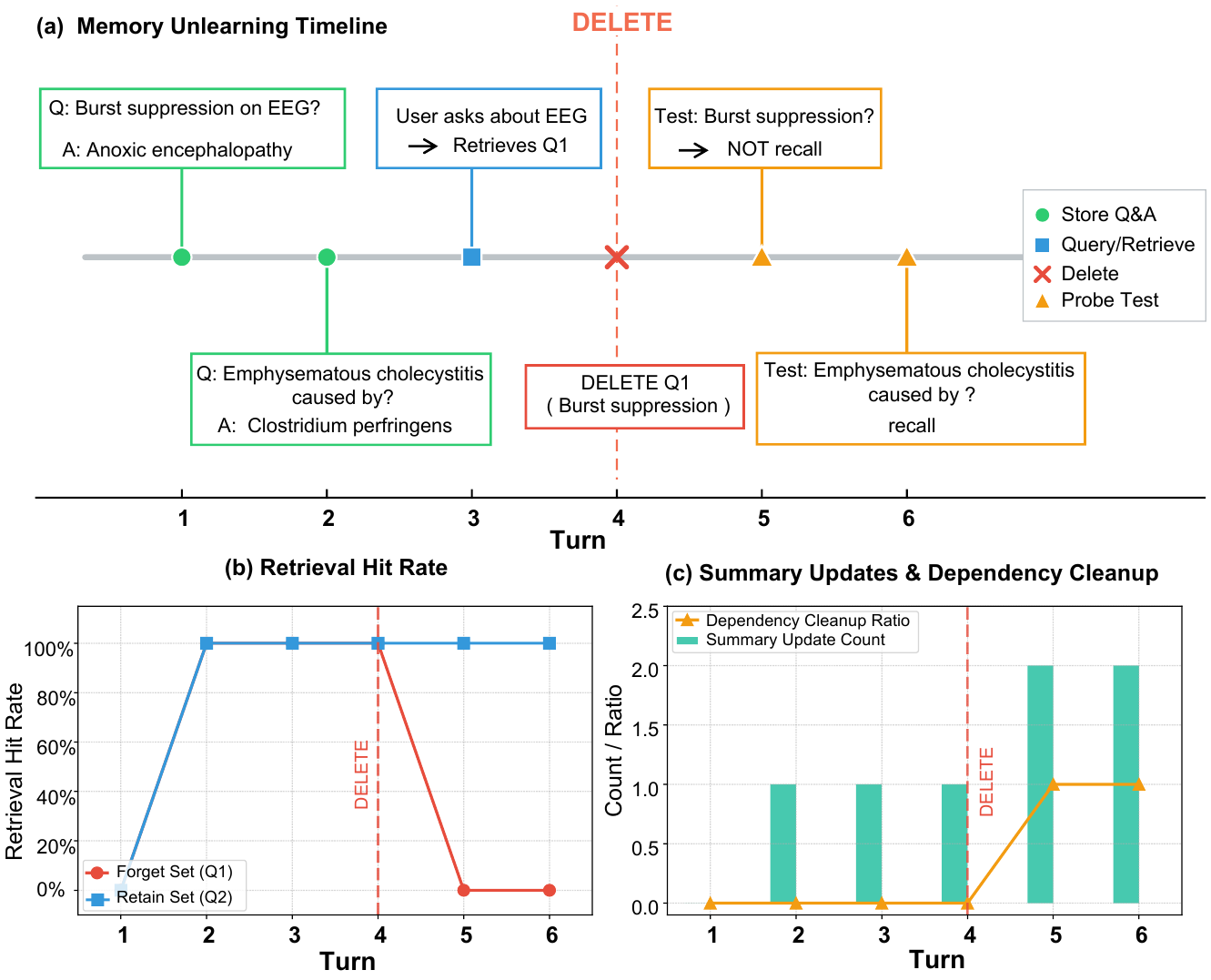}
\caption{Agent Loop evaluation. (a) Timeline. Q1 (Forget) is deleted at T4; Q2 (Retain) persists. (b) Retrieval hit rate. Forget drops to 0\% post-deletion; Retain is stable. (c) Summary updates and dependency cleanup ratio.}
\label{fig:agent-loop}
\end{figure}

\subsection{Ablation Studies}

\indent\textit{\textbf{Effect of the parameter and memory pathways.}}
We evaluate each pathway's contribution by comparing SBU with two variants: \texttt{w/o Mem} (parameter-level unlearning only) and \texttt{Mem-Only} (memory unlearning with frozen parameters). As shown in Table~\ref{tab:ablation-pathways}, removing the memory pathway leads to a noticeable degradation in MIA-based privacy metrics. In contrast, the \texttt{Mem-Only} variant underperforms on forget and test set accuracies, despite some privacy improvement. 
Ours suppresses the membership inference attack without sacrificing retain set accuracy, demonstrating that parameter updates and memory governance are complementary.

\indent\textit{\textbf{Hyperparameter sensitivity analysis.}}
We evaluate 34 configurations on MedMCQA to study the sensitivity of parameter unlearning to $\lambda_F$, temperature $T$, and entropy fallback, as visualized in Figure~\ref{fig:3d-surface}. The optimal configuration ($\lambda_F{=}1.5$, $T{=}2.0$, with entropy fallback) achieves 90\% test accuracy while reducing forget accuracy to 68\%, demonstrating effective knowledge removal without compromising utility. 
Our analysis reveals that entropy fallback is critical for preserving general capabilities, improving test accuracy from 78\% to 90\% and reducing forget accuracy from 80\% to 68\% under identical $\lambda_F$ and $T$ settings. 
The forgetting coefficient $\lambda_F$ controls forgetting-utility trade-off: values below 0.5 fail to induce sufficient forgetting (forget accuracy 86\%); excessive values ($\lambda_F{\geq}3.0$) cause test degradation (66\%) despite stronger forgetting (forget accuracy 64\%). Temperature exhibits a narrow effective range, with moderate values (1.0--2.0) performing best, whereas high temperature ($T{=}4.0$) destabilizes optimization and drops test accuracy to 68\%.

\indent\textit{\textbf{Memory-side unlearning analysis.}} 
Rebuilding the vector index improves privacy (MIA AUC: $0.5467 \rightarrow 0.5180$) but remains above the Retraining Oracle (0.5020) and the ideal 0.5 (Table~\ref{tab:memory-ablation}). Ours aligns with the Oracle in both utility ($0.9960$ vs. $0.9920$) and privacy verification (MIA AUC: $0.5000$ vs. $0.5020$), without system reconstruction.

\indent\textit{\textbf{Agent Loop evaluation.}} 
To demonstrate unlearning within an interactive loop, rather than a one-time procedure on a static dataset, we design an Agent Loop evaluation (Figure~\ref{fig:agent-loop}(a)) with four stages: Store, Query, Delete, and Probe. The agent stores medical Q\&A pairs (T1--T2), retrieves memories to answer user queries (T3), processes a deletion request (T4), and is probed for verification (T5--T6). As shown in Figure~\ref{fig:agent-loop}(b), the Forget Set retrieval hit rate drops from 100\% to 0\% while the Retain Set remains accessible. Figure~\ref{fig:agent-loop}(c) further confirms that summary updates and dependency cleanup prevent indirect leakage through derived content. Finally, a diagnostic case study (see the supplementary material) shows single-pathway unlearning can suffer from cross-pathway backflow, whereas our dual-pathway design suppresses backflow and enables genuine agentic unlearning.

\begin{table}[t]
\centering
\caption{Ablation Study: Memory-side Unlearning Strategies on MedMCQA (Forget Size = 100)}
\label{tab:memory-ablation}
\resizebox{\columnwidth}{!}{%
\begin{tabular}{lccccc}
\toprule
\textbf{Method} & \textbf{Forget Acc $\downarrow$} & \textbf{Retain Acc $\uparrow$} & \textbf{Test Acc $\uparrow$} & \textbf{MIA AUC $\rightarrow$0.5} & \textbf{MIA Score $\uparrow$} \\
\midrule
Naïve Deletion &
$88.33_{\pm 2.08}$ & $88.00_{\pm 1.00}$ & $89.67_{\pm 1.15}$ &
$0.5467_{\pm 0.0150}$ & $0.9067_{\pm 0.0300}$ \\
Re-indexing &
$82.67_{\pm 3.51}$ & $88.33_{\pm 1.53}$ & $89.33_{\pm 1.53}$ &
$0.5180_{\pm 0.0120}$ & $0.9640_{\pm 0.0240}$ \\
Retraining Oracle &
$\underline{72.00_{\pm 4.00}}$ & $\underline{89.00_{\pm 1.00}}$ & $\underline{89.67_{\pm 0.58}}$ &
$\underline{0.5020_{\pm 0.0080}}$ & $\underline{0.9920_{\pm 0.0160}}$ \\
\midrule
\rowcolor{gray!10}
\textbf{Ours} &
$\mathbf{71.33_{\pm 7.02}}$ & $\mathbf{89.33_{\pm 1.53}}$ & $89.00_{\pm 1.00}$ &
$\mathbf{0.5000_{\pm 0.0265}}$ & $\mathbf{0.9960_{\pm 0.0020}}$ \\
\bottomrule
\end{tabular}%
}
\end{table}

\begin{table}[ht!]
\centering
\footnotesize
\caption{Ablation study on parameter and memory pathways.}
\label{tab:ablation-pathways}
\setlength{\tabcolsep}{3pt}
\begin{tabular}{@{}lccccc@{}}
\toprule
Method & Forget $\downarrow$ & Test $\uparrow$ & Gen. $\uparrow$ & MIA AUC & MIA Score $\uparrow$ \\
\midrule
w/o Mem & 74.33 & 91.00 & 89.00 & 0.583 & 0.834 \\
Mem-Only & 85.67 & 88.33 & 86.67 & 0.621 & 0.758 \\
\rowcolor{gray!10}
\textbf{Ours} & \textbf{73.00} & \textbf{92.50} & \textbf{90.50} & \textbf{0.552} & \textbf{0.895} \\
\bottomrule
\end{tabular}
\end{table}

\section{Conclusion}
\label{sec1}
We study agentic unlearning for LLM agents with external memory, and identify parameter-memory backflow: a recontamination loop between model weights and retrieved data. We propose the Synchronized Backflow Unlearning (SBU) framework that integrates parameter and memory unlearning via a synchronized dual-pathway protocol. The parameter path optimizes a KL divergence objective to guide outputs to a high-entropy prior. The memory path uses dependency closure to prune isolated data while logically invalidating shared artifacts. Together, they form a closed-loop system preventing re-activation. Experiments on medical QA datasets demonstrate that SBU outperforms existing baselines in forgetting private information across both pathways, while preserving high fidelity on the retention set and maintaining computational overhead.
A limitation of our current approach is that dependency tracking may not fully capture cross-agent information flow in shared knowledge graphs. Future work will explore unlearning protocols specifically tailored for multi-agent collaborative environments.

\section*{Ethical Statement}

This work uses publicly available medical QA datasets (MedQA, MedMCQA, MedReason) that do not contain real patient identifiers. No real patient data was collected or processed. Deployment in clinical settings would require additional regulatory review and institutional approval.


\bibliographystyle{named}
\bibliography{ijcai26}

\end{document}